\pdfoutput=1

\documentclass[11pt]{article}

\usepackage{EACL2023}

\usepackage{times}
\usepackage{latexsym}

\usepackage[T1]{fontenc}

\usepackage[utf8]{inputenc}

\usepackage{microtype}

\usepackage{inconsolata}

\usepackage{graphics}
\usepackage{enumitem}
\usepackage{booktabs}
\usepackage{todonotes}
\usepackage{hyperref}
\usepackage{listings}
\usepackage{xcolor}
\newcommand{\squiggle}{\char`~}

\definecolor{codegreen}{rgb}{0,0.6,0}
\definecolor{codegray}{rgb}{0.5,0.5,0.5}
\definecolor{codepurple}{rgb}{0.58,0,0.82}
\definecolor{blk}{rgb}{0,0,0}
\definecolor{backcolour}{rgb}{0.95,0.95,0.95}

\lstdefinestyle{mystyle}{
    backgroundcolor=\color{backcolour},
    commentstyle=\color{codegray},
    stringstyle=\color{blk},
    basicstyle=\ttfamily\footnotesize,
    breakatwhitespace=false,         
    breaklines=true,                 
    captionpos=b,                    
    keepspaces=true,                 
    numbersep=5pt,                  
    showspaces=false,                
    showstringspaces=false,
    showtabs=false,                  
    tabsize=2
}
\title{Aligning the Norwegian UD Treebank with Entity and Coreference Information}

\author{Tollef Emil Jørgensen \\
  Norwegian University of Science and Technology \\
  \texttt{tollef.jorgensen@ntnu.no} \\\And
  Andre Kåsen \\
  National Library of Norway \\
  \texttt{andre.kasen@nb.no} \\}

\begin{document}
\maketitle
\begin{abstract}
This paper presents a merged collection of entity and coreference annotated data grounded in the Universal Dependencies (UD) treebanks for the two written forms of Norwegian: Bokmål and Nynorsk. The aligned and converted corpora are the \textit{Norwegian Named Entities} (NorNE) and \textit{Norwegian Anaphora Resolution Corpus} (NARC). While NorNE is aligned with an older version of the treebank, NARC is misaligned and requires extensive transformation from the original annotations to the UD structure and CoNLL-U format.
We here demonstrate the conversion and alignment processes, along with an analysis of discovered issues and errors in the data -- some of which include data split overlaps in the original treebank.
These procedures and the developed system may prove helpful for future corpus alignment and coreference annotation endeavors.
The merged corpora comprise the first Norwegian UD treebank enriched with named entities and coreference information.
\end{abstract}

\section{Introduction}
Resources for the Norwegian language have drastically increased in the last few years. Large text corpora such as the Norwegian Newspapers Corpus\footnote{\url{https://www.nb.no/sprakbanken/ressurskatalog/oai-nb-no-sbr-4/}} and the Norwegian Colossal Corpus \citep{kummervold-etal-2022-norwegian} supported the development of transformer-based models: \textit{NB-BERT} \citep{kummervold-etal-2021-operationalizing} and \textit{NorBERT} \citep{kutozov21}. Moreover, there are task-specific resources for document-level and fine-grained sentiment analysis \citep{VelOvrBer18, barnes-etal-2019-lexicon, OvrMaeBar20}, dependency syntax, part-of-speech, morphological features, lemmatization \citep{solberg-etal-2014-norwegian,ovrelid2016universal}, named entity recognition \citep{norne} and coreference resolution \citep{maehlum-etal-2022-narc}.
\\
\\
In addition to UD Norwegian Bokmål and UD Norwegian Nynorsk, there are two more available treebanks: 1) \textit{Language Infrastructure made Accessible} (LIA) \citep{ovrelid-etal-2018-lia} and 2) \textit{Norwegian Dialect Corpus} (NDC) \citep{kaasen2022norwegian}. These are based on speech transcripts rather than written sources like the former two. LIA is also converted to UD with the procedure from \citet{ovrelid2016universal}.


Currently, no up-to-date baselines\footnote{There is, however, an earlier effort for Norwegian coreference found in: \citet{borthen2004annotation}, \citet{noklestad2006detecting}, \citet{holen2007automatic}, \citet{johansson2008improving} and \citet{noklestad2009machine}} exist for Norwegian coreference resolution, which motivated this work in part of conforming to the CorefUD initiative \citep{nedoluzhko-etal-2022-corefud}, with the goal of unifying coreference corpora to a standardized CoNLL-U format.

The following sections describe related work, an overview of data sources and statistics, conversion, alignment with UD, error analysis, conclusions, and limitations.

\section{Related Work}
NARC is annotated using the BRAT annotation tool \citep{stenetorp2012brat}.
While conversion scripts are available for the resulting pairs of \textit{.ann} and \textit{.txt} files, such as the official from BRAT\footnote{\url{https://github.com/nlplab/brat/tree/master/tools}}, none sufficed for the annotation scheme used in NARC, due to cases like discontinuous mentions, validation checks for self-referring clusters and more.
We can find an example of BRAT outputs and CoNLL in the Litbank corpus \citep{litbank}, but the initial annotations used in BRAT are unlike NARC, nor is there available code.
We set up a conversion pipeline to the commonly used JSON line format for coreference resolution, as popularized by \citet{lee-etal-2018-higher}, and finally to CoNLL-U\footnote{\url{https://universaldependencies.org/format.html}}, conforming to the CorefUD standards and validation requirements \citep{nedoluzhko-etal-2022-corefud}. The procedures were validated throughout the alignment process using tools from UD\footnote{\url{https://github.com/UniversalDependencies/tools}} and Udapi \citep{popel2017udapi}.


\section{Data}
Three key data sources are involved in this project: UD treebanks for Bokmål and Norwegian, NARC, and NorNE. Following are brief descriptions along with statistics on the merging process.
\subsection{Universal Dependencies}
The current UD treebank is based on the Norwegian treebank \citep{solberg-etal-2014-norwegian}, one of the first widely used resources for Norwegian, initially developed within an in-house framework corresponding to the theories and practices described and documented in \cite{faarlund1998norsk}. The inventory of part-of-speech tags follows those defined for the Oslo-Bergen tagger \cite{hagen2000constraint}.
\par
The treebank was later converted and included in Universal Dependencies \citep{ovrelid2016universal}. It is structured in the CoNLL-U format, bound by sentence identifiers without document-level bounds, as shown in Appendix \ref{app:udbmdev}.
As of April 2023, the UD treebank for both Bokmål\footnote{\url{https://github.com/UniversalDependencies/UD\_Norwegian-Bokmaal\#changelog}} and Nynorsk\footnote{\url{https://github.com/UniversalDependencies/UD\_Norwegian-Nynorsk\#changelog}} have been updated to the latest version of UD (version 2.12).


\subsection{NARC}
NARC \citep{maehlum-etal-2022-narc} is the first openly available corpus for Norwegian coreference resolution.
The corpus consists mainly of news texts (85\%), the rest being government reports, parliamentary transcripts, and blog posts. Its annotations include markables, either as singleton mentions or as referred relational mentions, the latter subdivided into the four types: anaphoric, cataphoric, split antecedent and bridging relations.
There are three major issues regarding conversion:
1) NARC is released per document, lacking sentence identifiers for direct alignment with UD. 2) It is annotated on a character-level basis, where the CoNLL-U format requires word-level annotations. 3) Some documents do not exist in the UD treebanks.
We will revisit the issues in section \ref{sec:corefconv}.

\subsection{NorNE}
NorNE \citep{norne} is one of the most extensive corpus for Norwegian named entities, annotated with persons, organizations, locations, geo-political entities, products, and events, in addition to a separate \textit{derived} class for nominals derived from a name. While the NorNE corpus is already an enrichment of the UD treebank, UD has since received updates, mostly in terms of corrected token HEADs. The alignment process only included extracting the CoNLL-U \textit{MISC} field (the named entities) from NorNE, placing them with their matching token indices in UD. For an experimental exploration of NorNE, the reader is advised to consult \citet{aasmoe2019named}. Earlier efforts for Norwegian with respect to NER can be found in both \citet{johannessen2005named}, \citet{haaland2008maximum} and \citet{johansen2019named}. The mentioned update of UD ensures NorNe, through the conversion processes described in this paper, inherits all updated values.

\subsection*{Statistics}
As annotated documents in NARC contain a subset of the existing UD documents, there is an obvious information loss. Full statistics on the number of sentences, tokens and more, across UD, NorNE and NARC can be found in Appendix \ref{app:stats}. The information loss from NARC, to the aligned final corpora, is shown in Table \ref{tab:statsloss}. We cannot reduce these losses, as the texts simply do not occur in UD. However, much of the lost data were unrelated terms preceding the document; an example of this is shown in Appendix \ref{app:narcex1}.

\begin{table}[!htbp]
\centering
\resizebox{\columnwidth}{!}{%
\begin{tabular}{@{}lrrr@{}}
\toprule
\multicolumn{1}{c}{\begin{tabular}[c]{@{}c@{}}NARC\\ Alignment\\ loss\end{tabular}} & \multicolumn{1}{c}{\begin{tabular}[c]{@{}c@{}}Bokmål\\ (\%)\end{tabular}} & \multicolumn{1}{c}{\begin{tabular}[c]{@{}c@{}}Nynorsk\\ (\%)\end{tabular}} & \multicolumn{1}{c}{Total} \\ \midrule
Sentences                                                                           & 789 (4.8\%)                                                               & 281 (2.2\%)                                                                & 1,070                     \\
Tokens                                                                              & 13,510 (5.2\%)                                                            & 6,562 (3.1\%)                                                              & 20,073                    \\
Markables                                                                           & 2,410 (4.4\%)                                                             & 1,071 (2.3\%)                                                              & 3,483                     \\
Mentions                                                                            & 3,582 (4.6\%)                                                             & 1,522 (2.4\%)                                                              & 5,104                     \\
\begin{tabular}[c]{@{}l@{}}SplitAnte\\ Clusters\end{tabular}                        & 6 (4.3\%)                                                                 & 1 (1.2\%)                                                                  & 7                         \\
\begin{tabular}[c]{@{}l@{}}Bridging\\ Clusters\end{tabular}                         & 35 (3.4\%)                                                                & 27 (3.1\%)                                                                 & 62                        \\ \bottomrule
\end{tabular}%
}
\caption{Information loss during the alignment of NARC}
\label{tab:statsloss}
\end{table}
\clearpage
We remind the reader that the corpus contains \squiggle85\% news texts, which often include topics, categories, and other text that may not be related to the article's main body. As such, the raw numbers may not represent an equal loss regarding usability and realistic use cases.

All numbers are extracted using Udapi \citep{popel2017udapi}, both its command-line tool and the Python integration\footnote{\url{https://github.com/udapi/udapi-python}} (\texttt{corefud.MiscStats} and \texttt{corefud.Stats} modules).
The \textit{NARC}-column represents converted CoNLL-U formatted NARC, whereas the \textit{Aligned}-column represents the aligned train/test/dev splits.
While the statistics differ from those presented in the original paper \citep{maehlum-etal-2022-narc}, the categories are described as follows:
\begin{itemize}
    \item Markables are all unique entities in the document (including singletons)
    \item Mentions are all occurrences and references to the markables
    \item Bridging- and split antecedent clusters refer to the count of grouped clusters of each respective mention type -- not the number of relations within each group.
\end{itemize}
See Appendix \ref{app:statsexpl} for examples.


\section{Coreference Conversion and Alignment}
\label{sec:corefconv}
The initial part of aligning NARC is converting the original annotation files (\textit{.ann/.txt} pairs) to the CoNLL-U format. A natural step along the way was to parse these files into the JSON line format with sentence, token, and clustering information. The JSON line files are converted to CoNLL-U and aligned with the UD treebanks.
\par
The steps involved are:
\subsubsection*{Ann$\rightarrow$JSON conversion}
\begin{enumerate}[label=(\alph*)]
\setlength\itemsep{0em}
    \item Extract markables and mentions, bridging and split antecedents, group discontinuous mentions
    \item Find connected clusters by building a graph of coreference links
    \item Map character-based indices to word indices
    \item Restructure word-indexed markables and clusters into a JSON line (one .jsonl per .ann)
\end{enumerate}

\subsubsection*{JSON$\rightarrow$CoNLL-U conversion}
\begin{enumerate}[label=(\alph*)]
\setlength\itemsep{0em}
    \item Adjust markables spanning tokens not in their equivalent UD spans
    \item Iteratively add markables and mention clusters token-wise, ensuring correct ordering of multi-entity spans according to UD standards (see UDs \textit{Level 6} validation for coreference and named entities\footnote{\url{https://github.com/UniversalDependencies/tools/blob/master/validate.py\#L2112}})
    \item Restructure according to the CoNLL-U format guidelines, populating the MISC column, leaving out empty fields to be filled by the UD treebank.
\end{enumerate}

\subsubsection*{NARC $\rightarrow$ UD alignment}
A highly compressed overview of the alignment process can be described as follows:
\begin{enumerate}[label=(\alph*)]
\setlength\itemsep{0em}
\setlength{\abovedisplayskip}{4pt}
\setlength{\belowdisplayskip}{4pt}
    \item Map UD sentence text $\rightarrow$ UD index
    \item Map UD index $\rightarrow$ train/test/dev split
    \item Process NARC documents and extract UD index candidate sentences (one-to-many)
    \item For every sentence with multiple candidates, extract its sentence identifiers in both NARC ($N$) and UD ($U$) and build a cost matrix based on the distances to neighboring indices:
    $$C_{i,j} = \texttt{sent\_to\_UD\_dist\_score}(N_i, U_j)$$
    We then disambiguate by minimizing sentence distances by solving the linear assignment problem for $C$ \citep{jonker1988shortest}.
    \item Verify whether a sentence index is part of more than one UD split. If so, discard the document.
   
\end{enumerate}

\section{Analysis}
\label{sec:error}
We discovered several issues and error patterns in the conversion and alignment processes -- some already mentioned in the steps above. The following error analysis documents problems with the current corpora and illustrates how the developed system may aid future alignment tasks in detecting errors, especially if one has a corpus managed and annotated by multiple parties.

\subsubsection*{Sentence mismatch and tokenization issues}
A typical error in NARC is an inserted pipe character (|) preceding \textit{some} commas and the end-of-sentences, which is not the case for UD data.
The extra character is often included in involved markable spans, and its end-index must be decremented accordingly.
A total of 2057 spans were corrected for 561 documents.
Another issue is two aligned sentences having different tokens (see Table \ref{tab:illust}).
In this case, we map 1:1 sentences to the UD tokens.
In the same analysis, four documents in NARC Bokmål (\texttt{klassekampen\_\{01,02,03,04\}}) were not found in UD Bokmål, but had matches in UD Nynorsk and should thus be moved.
\begin{table}[!htbp]
\centering
\setlength\belowcaptionskip{-15pt}
\begin{tabular}{@{}lll@{}}
\toprule
NARC sentence & UD sentence \\ \midrule
Illustrasjonsfoto . & Illustrasjonsfoto \\
Illustrasjonsfoto | & Illustrasjonsfoto \\
Illustrasjonsofoto | & Illustrasjonsfoto . \\
Nei ! & - Nei ? \\
Nei ! & - Nei . \\
- Ja . & Ja . \\
\bottomrule
\end{tabular}
\caption{Examples of tokenization mismatch}
\label{tab:illust}
\end{table}

\subsubsection*{Duplicates and multiple sentence matches}
Most commonly occurring in dialogue-based texts, we may observe recurring sentences like ``illustrasjonsfoto'' (illustration photo), ``les også'' (read also), interjections, and entity names included multiple times throughout a document. Pure string matching would fail in these cases, such as in the following example, where two people (\textit{Elling} and \textit{Espen}) have several mentions in a dialogue setting. The numbers are sentence indexes where the sentence itself is either \textit{Elling} or \textit{Espen}.
\begin{lstlisting}[belowskip=-0.5\baselineskip]
'Elling': [15, 26, 41, 56, 63, 79, 87, 97, 103, 108, 114, 119],
'Espen': [33, 45, 65, 74, 91, 99, 106, 110, 117]
\end{lstlisting}
\begin{center}
    \fontsize{9}{9}
    \textit{Example 1: Elling and Espen mentioned in a dialogue setting (doc: kknn\squiggle20030124-27894)}
\end{center}
There are, in total, 597 ambiguous sentences across 234 documents. These are resolved by the sentence disambiguation process in step (d) above.
\subsubsection*{Lemma injection}
In rare cases, sentences have no symmetric match (even after preprocessing for tokenization issues) in both NARC and UD. Two of these were found to have a lemma injected in place of their original entry.
\begin{enumerate}
    \setlength\itemsep{0em}
    \item \texttt{vtbnn\squiggle20090625-4275}, sentence 23. ``kostar \textbf{vi} mykje" (costs we a lot) where \textbf{vi} (we) is \textbf{oss} (us) in UD Nynorsk test, ID 017342.
    \item \texttt{firdann\squiggle20100305-5007021}, sentence 15. ``ordførar'' (mayor) is ``ordføraren'' (the mayor) in UD Nynorsk train, ID 005311.
\end{enumerate}
\texttt{vtbnn\squiggle20031111-1592} has a unique error, where the conjunction ``at'' (that) is in place of the adposition ``ved'' (by), token 26 of UD Nynorsk train, ID 012440.

\subsubsection*{Data split overlap}
Eleven documents were found to span train/test/dev splits in the original treebanks (6 Bokmål, 5 Nynorsk). Although comprising one coherent text, these documents have two parts (with no logical separation), each in a different split in UD. The suggested correction is to update the original treebanks to contain the entire document. Details are found in Appendix \ref{app:udsplit}.

\section{Limitations}
While the system may be applied to other UD-related expansions, task specific details must be customized in the pipeline. Further, there are likely more UD alignment errors to uncover for data sources other than those described here.
\section{Conclusions}
We have presented the merging and alignment of NARC, NorNE, and UD for Norwegian Bokmål and Nynorsk, along with statistics of the final corpora. The processes are modular in the sense that updates to any of the corpora will be supported and will still align with their root in UD. With the developed system supporting the conversion of BRAT annotation files and the alignment of treebanks, we have been able to maximize the included data throughout the merging process. Future work is twofold: 1) correct the data split overlaps in UD and 2) adjust the NARC annotation files according to the findings here to avoid future errors. All related code can be found in the repository UD-NARC\footnote{\url{https://github.com/tollefj/UD-NARC}}.

\section*{Acknowledgements}
Thanks to Michal Novák and Daniel Zeman for valuable feedback throughout the conversion and alignment process.
\bibliography{custom}
\bibliographystyle{acl_natbib}
\newpage
\appendix
\section*{Appendices}
\label{sec:appendix}
\section{Annotation Mismatch}
\subsection{Excerpt from UD Bokmål (dev)}
\label{app:udbmdev}
First two sentences of \textit{no\_bokmaal-ud-dev.conllu}. Tokens after index 2 are replaced with $\cdots$.

The first sentence aligns with line number 19 in the NARC dataset of the corresponding file: \texttt{ap\squiggle20091016-3323000.txt}, shown in Appendix \ref{app:narcex1}

\begin{lstlisting}
# sent_id = 015697
# text = Dommer Finn Eilertsen avstår, selvfølgelig bevisst, fra å "sette ord på" det inntrykk retten for sitt vedkommende måtte ha dannet seg av de handlinger retten finner bevist og av lovovertrederen.
1	Dommer	dommer	NOUN	_	Definite=Ind|Gender=Masc|Number=Sing	2	nmod	_	_
2	Finn	Finn	PROPN	_	Gender=Masc	4	nsubj	_	_
...
# sent_id = 015698
# text = Dommeren lar gjerningsbeskrivelsen tale for seg uten karakteristikk og uten å ty til de moralsk fordømmende ord.
1	Dommeren	dommer	NOUN	_	Definite=Def|Gender=Masc|Number=Sing	2	nsubj	_	_
2	lar	la	VERB	_	Mood=Ind|Tense=Pres|VerbForm=Fin	0	root	_	_
...
\end{lstlisting}

\subsection{Excerpt from NARC}
\label{app:narcex1}
First 20 lines of \texttt{ap\squiggle20091016-3323000.txt}. Note that we first start matching with the equivalent UD-source on line 19. Several noun-phrases preceding this line are annotated in NARC.
\begin{lstlisting}[numbers=left]
Det notoriske rovdyr |
Serievoldtekt .
Ikke til å leve med .
Et skille .
En ønsket utvikling ?
Et spark .
Smitteeffekt .
Slått ut .
Begravende journalistikk .
Sosialarbeider .
Respekt som menneske .
Aftenpostens intervju med en meddommer i samvittighetskval er ikke gravende journalistikk , det er begravende journalistikk .
Oslo tingrett avsa fredag 9. oktober dom i den såkalte serievoldtektssak som Aftenposten har dekket .
Dommen er ikke rettskraftig .
Jeg går ikke inn på dommens materielle innhold , dvs. rettens vurderinger av skyldspørsmål og vedrørende straffeutmåling .
I formell henseende fremstår dommen som uklanderlig :
Logisk og enkelt bygget opp , ryddig , lettlest .
Og med et tindrende særpreg :
Dommer Finn Eilertsen avstår , selvfølgelig bevisst , fra å " sette ord på " det inntrykk retten for sitt vedkommende måtte ha dannet seg av de handlinger retten finner bevist og av lovovertrederen .
Dommeren lar gjerningsbeskrivelsen tale for seg uten karakteristikk og uten å ty til de moralsk fordømmende ord .
\end{lstlisting}

\subsubsection*{Corresponding annotations}
\label{app:bad-annotations}
Following the mismatch between UD and NARC, annotations up until T$37$ (line 37) are invalid entities in the merged data. The first 10 markables do not relate to the news article body.
\begin{lstlisting}[numbers=left]
T1	Markable 0 22	Det notoriske rovdyr |
T2	Markable 23 36	Serievoldtekt
T3	Markable 61 72	Et skille .
T4	Markable 73 94	En ønsket utvikling ?
T5	Markable 95 105	Et spark .
T6	Markable 106 120	Smitteeffekt .
T7	Markable 132 158	Begravende journalistikk .
T8	Markable 159 175	Sosialarbeider .
T9	Markable 176 198	Respekt som menneske .
T10	Markable 188 196	menneske
T11	Markable 199 211	Aftenpostens
T12	Markable 199 257	Aftenpostens intervju med en meddommer i samvittighetskval
T13	Markable 225 257	en meddommer i samvittighetskval
T14	Markable 240 257	samvittighetskval
T15	Markable 266 288	gravende journalistikk
T16	Markable 291 294	det
T17	Markable 298 322	begravende journalistikk
T18	Markable 325 338	Oslo tingrett
T19	Markable 344 361	fredag 9. oktober
T20	Markable 351 361	9. oktober
T21	Markable 362 365	dom
T22	Markable 368 424	den såkalte serievoldtektssak som Aftenposten har dekket
T23	Markable 402 413	Aftenposten
T24	Markable 427 433	Dommen
T25	Markable 457 460	Jeg
T26	Markable 477 484	dommens
T27	Markable 477 577	dommens materielle innhold , dvs. rettens vurderinger av skyldspørsmål og vedrørende straffeutmåling
T28	Markable 511 518	rettens
T29	Markable 506 577	dvs. rettens vurderinger av skyldspørsmål og vedrørende straffeutmåling
T30	Markable 506 547	dvs. rettens vurderinger av skyldspørsmål
T31	Markable 534 547	skyldspørsmål
T32	Markable 562 577	straffeutmåling
T33	Markable 582 599	formell henseende
T34	Markable 609 615	dommen
T35	Markable 620 631	uklanderlig
T36	Markable 691 711	et tindrende særpreg
T37	Markable 714 735	Dommer Finn Eilertsen
\end{lstlisting}

\section{Statistics}
\label{app:stats}
Tables \ref{tab:stats_bm} and \ref{tab:statsnn} show the detailed numbers for sentences, tokens, entities, markables, mentions, split antecedent clusters and bridging clusters.

\begin{table}[!htbp]
\centering
\resizebox{\columnwidth}{!}{%
\begin{tabular}{@{}lrrrr@{}}
\toprule
Bokmål                                                      & \multicolumn{1}{c}{UD} & \multicolumn{1}{c}{NorNE} & \multicolumn{1}{c}{NARC} & \multicolumn{1}{c}{Aligned} \\ \midrule
Sentences                                                    & 20,044                 & 20,045                    & 16,461                   & 15,672                                                                                    \\ \midrule
Tokens                                                       & 310,221                & 310,222                   & 257,646                  & 244,136                                                                                   \\ \midrule
Entities                                                     & -                      & 20,134                    & -                        & 16,271                                                                                    \\ \midrule
Markables                                                    & -                      & -                         & 55,225                   & 52,815                                                                                    \\ \midrule
Mentions                                                     & -                      & -                         & 77,565                   & 73,983                                                                                    \\ \midrule
\begin{tabular}[c]{@{}l@{}}SplitAnte\\ Clusters\end{tabular} & -                      & -                         & 140                      & 134                                                                                       \\ \midrule
\begin{tabular}[c]{@{}l@{}}Bridging\\ Clusters\end{tabular}  & -                      & -                         & 1,060                    & 1,025                                                                                     \\ \bottomrule
\end{tabular}%
}
\caption{Statistics of the Bokmål corpora}
\label{tab:stats_bm}
\end{table}

\begin{table}[!htbp]
\centering
\resizebox{\columnwidth}{!}{%
\begin{tabular}{@{}lrrrr@{}}
\toprule
Nynorsk                                                      & \multicolumn{1}{c}{UD} & \multicolumn{1}{c}{NorNE} & \multicolumn{1}{c}{NARC} & \multicolumn{1}{c}{Aligned} \\ \midrule
Sentences                                                    & 17,575                 & 17,575                    & 12,762                    & 12,481                      \\ \midrule
Tokens                                                       & 301,353                & 301,353                   & 213,222                  & 206,660                     \\ \midrule
Entities                                                     & -                      & 20,087                    & -                        & 15,520                      \\ \midrule
Markables                                                    & -                      & -                         & 45,918                   & 44,847                      \\ \midrule
Mentions                                                     & -                      & -                         & 63,137                   & 61,615                      \\ \midrule
\begin{tabular}[c]{@{}l@{}}SplitAnte\\ Clusters\end{tabular} & -                      & -                         & 81                       & 80                          \\ \midrule
\begin{tabular}[c]{@{}l@{}}Bridging\\ Clusters\end{tabular}  & -                      & -                         & 868                      & 841                         \\ \bottomrule
\end{tabular}%
}
\caption{Statistics of the Nynorsk corpora}
\label{tab:statsnn}
\end{table}
\clearpage

\subsection{Naming}
\label{app:statsexpl}
The following examples illustrate how Markables, Mentions, Bridge clusters and Split antecedent clusters are counted. Only Token and MISC columns included.
\subsubsection*{Bridge Example}
\begin{itemize}
\setlength\itemsep{0em}
    \item Markables: 3
    \item Mentions: 5
    \item Bridge Clusters: 1
\end{itemize}
\begin{lstlisting}
Kidnapperne	Entity=(1)
kom	_
seg	Entity=(1)
senere	_
unna	_
fordi	_
kystvakten	Entity=(2)
var	_
redd	_
de	Entity=(1)
ville	_
senke	_
båten	Bridge=2<3|Entity=(3)
.	_
\end{lstlisting}
\subsection*{Split Antecedent Example}
\begin{itemize}
\setlength\itemsep{0em}
    \item Markables: 6
    \item Mentions: 6
    \item SplitAnte clusters: 1 (only one cluster, but two \textit{mentions} within the cluster)
\end{itemize}
\begin{lstlisting}
Hennes	Entity=(1(2)
fraseparerte	_
ektemann	SpaceAfter=No|name=O
,	_
som	_
har	_
hentet	_
barnet	Entity=(3
deres	SplitAnte=1<4,2<4|Entity=(4)3)
noen	Entity=(5
dager	Entity=5)
tidligere	SpaceAfter=No|Entity=1)
,	_
er	_
ikke	_
å	_
få	_
tak	Entity=(6
i	SpaceAfter=No|Entity=6)
.	_
\end{lstlisting}

\newpage
\section{Universal Dependencies Data Split Overlap}
\label{app:udsplit}
\begin{table}[!htbp]
\resizebox{\columnwidth}{!}{%
\begin{tabular}{@{}lccc@{}}
\toprule
Document & Train & Test & Dev \\ \midrule
ap\squiggle 20081210-2445517 (Bokmål) &  & x & x \\
ap\squiggle 20091016-3323000 (Bokmål & x &  & x \\
bt\squiggle BT-20120916-2765289b (Bokmål) &  & x & x \\
db\squiggle 20081128-3858534b (Bokmål) &  & x & x \\
kk\squiggle 20110829-59221 (Bokmål) &  & x & x \\
vg\squiggle VG-20121219-10048819 (Nynorsk) &  & x & x \\
firdann\squiggle 20100118-4812178 (Nynorsk) &  & x & x \\
firdann\squiggle 20110916-5739806 (Nynorsk) & x &  & x \\
kknn\squiggle 20030804-23304 (Nynorsk) &  & x & x \\
vtbnn\squiggle 20070403-3233 (Nynorsk) & x &  & x \\
vtbnn\squiggle 20090625-4275 (Nynorsk) &  & x & x \\ \bottomrule
\end{tabular}%
}
\caption{Documents with parts corresponding to multiple data splits in the Universal Dependencies treebanks.}
\label{tab:splits}
\end{table}

\end{document}